\documentclass{article}

\usepackage{arxiv}
\usepackage{xcolor,soul,framed} 

\usepackage[utf8]{inputenc} 
\usepackage[T1]{fontenc}    

\usepackage{multirow}
\usepackage{lineno,hyperref}
\usepackage{url}            
\usepackage{cite}
\usepackage{natbib}
\usepackage{booktabs}       
\usepackage{amsfonts}       
\usepackage{amsmath}
\usepackage{amssymb}
\usepackage{nicefrac}       
\usepackage{microtype}      
\usepackage{color}          
\usepackage{theorem}
\usepackage{bm}
\usepackage{graphicx}
\usepackage{subfigure}
\usepackage{multicol}

\usepackage{nomencl}
\makenomenclature
\usepackage{etoolbox}
\renewcommand\nomgroup[1]{%
  \item[\bfseries
  \ifstrequal{#1}{A}{GP algorithm}{%
  \ifstrequal{#1}{B}{Online GP algorithm}{%
  \ifstrequal{#1}{C}{Online PAC-Bayes bound}{}}}%
]}

\usepackage{makecell}
\usepackage{algorithmicx}
\usepackage{algpseudocode}
\usepackage{algorithm,algpseudocode,float}

\allowdisplaybreaks[4]

\graphicspath{{figures/}} 

\usepackage{algpseudocode}
\usepackage{algorithm,algpseudocode,float}

\newtheorem{theorem}{Theorem}[section]

\newtheorem{corollary}[theorem]{Corollary}

\theoremstyle{definition}

\theoremstyle{remark}

\theoremstyle{assumption}

\theoremstyle{proof}

\title{Streaming  PAC-Bayes Gaussian process regression with a performance guarantee for online decision making
}

\author{Tianyu Liu \\
        \thanks{Australian Artificial Intelligence Institute, University of Technology Sydney. Australia. Email address: Tianyu.Liu-1@student.uts.edu.au, jie.lu@uts.edu.au, Yan.zheng@uts.edu.au, guangquan.zhang@uts.edu.au} \\ 
        \And Jie~Lu \\ \And Zheng~Yan \\ \And Guangquan Zhang     \\
}

\begin{document}
\maketitle

\begin{abstract}
As a powerful Bayesian non-parameterized algorithm, the Gaussian process (GP) has performed a significant role in Bayesian optimization and signal processing. GPs have also advanced online decision-making systems because their has a closed-form posterior distribution solution. However, its training and inference process requires all historic data to be stored and the GP model to be trained from scratch. For those reasons, several online GP algorithms have been specifically designed for streaming settings. In this work, we present a novel theoretical framework for online GPs based on the online probably approximately correct (PAC) Bayes theory. The framework offers both a guarantee of generalized performance and good accuracy. Based on this framework, an online PAC-Bayes GP algorithm (O-PACGP) with a bounded loss function is further proposed. This algorithm offers a balance between the generalization error upper bound and accuracy. Instead of minimizing the marginal likelihood, our O-PACGP algorithm directly minimizes the generalization error upper bound. In addition to its theoretical appeal, the algorithm performs well empirically on several regression datasets. Compared to other online GP algorithms, ours yields a generalization guarantee and very competitive accuracy.
\end{abstract}


\section{Introduction}\label{section_1}
GP is a powerful Bayesian non-parameterized algorithm that has an explicit closed-form posterior distribution. For this reason, it has played a significant role in both signal processing \cite{wang2007gaussian} and Bayesian optimization \cite{kandasamy2017query, nuara2022online}. It can both provide accuracy and estimation confidence. It not only provides accuracy but also high confidence in the estimation. However, training with standard GP comes at a high computational cost $O(N^3)$. Hence, sparse GPs, such as FITC \cite{snelson2005sparse}, VFE \cite{titsias2009variational}, were developed to reduce the time complexity. By introducing the $M$ ($M\ll N$) inducing points, not only is the time complexity is reduced from $O(N^3)$ down to $O(NM^2)$, but it also means these sparse GP algorithms can be scaled to large datasets.

However, these GPs cannot deal with streaming data, which means the data arrive sequentially in an online fashion in a series of small batches of an unknown number. In online settings, traditional GP requires that the classical or sparse GP algorithms incorporate the new streaming data into the old dataset. The posterior distribution is then retrained from scratch. This framework is optimal, but it requires that all historical data be stored. \cite{bui2017streaming} seeking a solution to reduce this space and time complexity, developed a new framework for GP learning and inference specifically for the streaming setting – a framework capable of online hyper-parameter learning and pseudo-point location optimization. Further, the solution involves a structured kernel interpolation approach to efficiently handle the online computations needed for constant-time $O(1)$ kernel hyper-parameters updates with respect to the number of points $n$ in \cite{stanton2021kernel}. Additionally, \cite{maddox2021conditioning} proposed a novel conditional SVGP algorithm the variational posterior is not required to be reoptimized through the evidence lower bound whenever new data is added.

Although several online GP methods have been proposed to deploy GP models in streaming settings, none provide a generalized guarantee of performance on future predictions with an unknown data distribution. In this paper, we propose a new online GP framework based on the online PAC-Bayes theory \cite{haddouche2022online} for online hyper-parameter learning and pseudo-point location optimization. Instead of maximizing the marginal likelihood, our algorithm optimizes the generalization error upper bound. This bound incorporates the empirical risk function and a regularization item, which is in proportion to the divergence between the prior distribution and posterior distribution of of the parameters. Compared to other online GP algorithms, our approach strikes a good balance between generalized performance and accuracy.

{\bf Problem setting} In this work, we focus on the streaming setting, which means the data arrive sequentially, but we have no knowledge of how much date will arrive in the interval. Additionally, the true distribution of the streaming data is also unknown. Our objective is to make predictions without need to store all the historical data, which means that the inducing points and the hyper-parameters must be updated in an online fashion when new streaming data is received to learn the model.

{\bf Our contributions} Although there are already several online GP algorithms specifically designed for streaming datasets \cite{bui2017streaming, stanton2021kernel, maddox2021conditioning}, none provide a guarantee of good generalized performance with an unknown data distribution. Motivated by the previous discussions, a new online PAC-Bayes bound based online GP framework is developed.

Thus the main contributions of this paper includes:

\begin{itemize}
\item In the online setting, traditional GPs requires all historic data to be stored, and the online GPs cannot provide generalization performance guarantee. Thus, a scalable online PAC-Bayes GP framework for streaming data is proposed which offers a quantified guarantee of generalization performance on future predictions given an unknown data distribution;
\item Based on the propose framework, an online PAC-Bayes GP algorithm (O-PACGP) with a bounded loss function is further investigated. The O-PACGP algorithm offers a balance between the generalization error upper bound and accuracy;
\item To verify the effectiveness of the proposed O-PACGP algorithm, several experiments with regression tasks datasets are conducted. The results illustrate that online PAC-Bayes GP methods achieve a competitive accuracy performance and superior generalization performance guarantee.
\end{itemize}

{\bf Outline} The rest of this paper is organized as follows. The literature review is provided in Section \ref{section_Related_work}, which covers GP and PAC-Bayes theory. Then, the standard GP/online GP, and online PAC-Bayes theory are introduced in Section \ref{section_Preliminaries}. Section \ref{section_online_pacgp} explains our novel online PACGP optimization approaches with bounded loss function for regression issue. Several experiments are provided to verify the O-PACGP algorithms in Section \ref{section_experiments}. Finally, Section \ref{section_conclusion} presents our conclusions.

\section{Related work}
\label{section_Related_work}
This section presents relevant work on online GP, PAC-Bayes theory, and the online PAC-Bayes approach.

\subsection{Online Gaussian process}
Online GP provides a probabilistic modeling paradigm to support decision-making in an online fashion. The first online GP framework was proposed by \cite{bui2017streaming}. It updates hyper-parameters and the induce points in a streaming way. The framework comprised two solutions: streaming sparse variational GPs (O-SVGP) and online sparse GP regression (O-SGPR). O-SGPR relies on closed-form marginalization and so is only suitable for Gaussian likelihoods. O-SVGP is applicable to non-Gaussian likelihoods; however, this approach sacrifices the closed-form expressions of the posterior distribution. Further, the solution incorporates a Woodbury inversion with structured kernel interpolation (WISKI) \cite{stanton2021kernel} to reduce the computational complexity. As such, it can perform online updates in constant-time $O(1)$, instead of $O(n)$, where $n$ is the number of new observations. In addition, \cite{maddox2021conditioning} develop Online Variational Conditioning (OVC), which is based on SVGP, to condition SVGPs on new data without reoptimizing the variational posterior through the evidence lower bound. This algorithm is also suitable for non-Gaussian likelihoods.

\subsection{PAC-Bayes theory}
 
\begin{figure}[tbp]
\centering 
\includegraphics[width=0.4\textwidth]{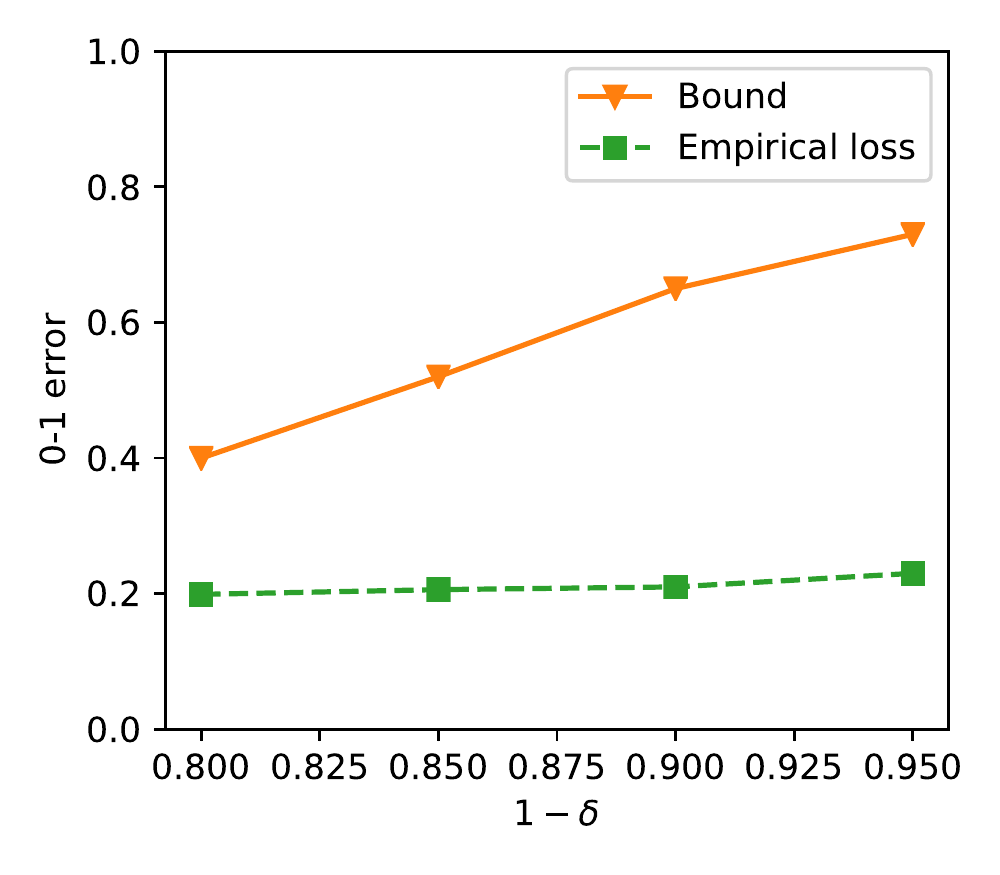}
\caption{With the confidence probability of $1-\delta$, the empirical error on the test dataset is lower than  error upper bound, which is estimated via PAC-Bayes theory. Here, the solid line is the estimated error upper bound via PAC-Bayes theory, and the dashed line indicates the empirical loss on the dataset. Besides, the higher confidence probability leads to a looser generalization error upper bound, and vice versa.
}
\label{PAC_Bayes_bound_new}
\end{figure}


PAC-Bayes bound provides a generalization performance guarantee for a learned model with selected optimization objective, which gives a probabilistic numerical upper bound. This bound has been applied in robot \cite{ren2021generalization}, stochastic neural network \cite{steffen2022pac, lyle2020benefits}, deterministic classifier \cite{clerico2022pac, biggs2022margins}, optimization (\cite{sucker2022pac, pitas2022cold, rivasplata2019pac, cherian2020efficient}), multi-view learning (\cite{sun2022stability}), and meta-learning (\cite{rezazadeh2022unified, liu2021statistical, nguyen2022pac, liu2021pac, chu2022dna}). The PAC-Bayes bound was first established by \cite{mcallester1999some}. However, this theory only applies to a bounded loss function and i.i.d data. So, \cite{germain2016pac} and \cite{alquier2016properties} developed PAC-Bayesian bounds for unbounded loss functions, i.e., sub-Gaussian and sub-Gamma loss families and negative log-likelihood function. In addition, \cite{haddouche2020pac} expanded PAC-Bayesian theory to learning problems with unbounded loss functions by introducing the condition of special boundedness. To achieve tighter PAC-Bayes bound, a novel conditionally Gaussian training algorithm that optimises the PAC-Bayesian bound is proposed in \cite{clerico2022conditionally}, without relying on any surrogate loss. Besides, \cite{biggs2022tighter} utilizes information about the difficulty of examples to obtain a tighter and fast-rate PAC-Bayesian generalisation bounds. \cite{grunwald2021pac} also extend PAC-Bayes theory to arbitrary VC classes by integrating mutual information and PAC-Bayes Bounds.

PAC-Bayes theory has also been used to optimize GP models operating in a batch setting. \cite{seeger2002pac} extend the PAC-Bayes theory to approximate Bayesian GP classification tasks and achieve tighter generalization error bounds. \cite{reeb2018learning} employed PAC-Bayes theory train a GP regression model, while \cite{liu2022robust} investigated robust PAC-Bayes noisy input GP algorithms. In addition, \cite{achituve2021personalized} derives a non-vacuous guarantees PAC-Bayes generalization bound for personalized federated Gaussian processes

However, those PAC-Bayes theories only hold in batch learning settings, which means that the entire dataset is accessible before inference. To extend this theory into the streaming setting, \cite{haddouche2022online} proposes an online PAC-Bayes learning framework. They developed two types of bounds: online PAC-Bayesian {\it training} and {\it test} bounds. The training bound exhibits the online procedures while the test bound provides the efficiency guarantees.

In this work, we deployed the online PAC-Bayes theory within a GP algorithm giving rise to an online PAC-Bayes GP algorithm that strikes a balance between the upper bound of the generalization error and accuracy.

\begin{table}[!htp]
\begin{framed}
\nomenclature[A,00]{\(n\)}{number of training data}
\nomenclature[A,01]{\(D\)}{dimension of training data}
\nomenclature[A,01]{\(X\)}{$n \times D$ training data matrix}
\nomenclature[A,02]{\(y\)}{$n \times 1$ training data vector}
\nomenclature[A,03]{\(X_\star\)}{$D$-dimensional test data matrix}
\nomenclature[A,04]{\(\epsilon_y\)}{output noise}
\nomenclature[A,06]{\(\sigma_y\)}{output noise variance}
\nomenclature[A,08]{\(f\)}{Gaussian process latent function values at training data}
\nomenclature[A,09.1]{\(\bar{f}\)}{{Gaussian process mean function}}
\nomenclature[A,09]{\(f_\star\)}{Gaussian process latent function values at testing data}
\nomenclature[A,10]{\(K\)}{$n\times n$ covariance matrix of training data $X$}
\nomenclature[A,11]{\(K_{f\star}\)}{covariance matrix between training data $X$ and testing data $X_\star$}
\nomenclature[A,12]{\(K_{\star\star}\)}{covariance matrix of $X_\star$}
\nomenclature[B,0]{\(y_{new}\)}{current data points}
\nomenclature[B,1]{\(y_{old}\)}{historical dataset}
\nomenclature[B,2]{\(\theta_{new}\)}{updated parameters after access to $y_{new}$ }
\nomenclature[B,3]{\(\theta_{old}\)}{previous approximation of true posterior}
\nomenclature[B,4]{\(q_{new}(f)\)}{new approximation of true posterior}
\nomenclature[B,5]{\(q_{old}(f)\)}{covariance matrix between $Z$ and $X_\star$}
\nomenclature[B,6]{\(p(f)\)}{exact posterior distribution}
\nomenclature[C,00]{\(S\)}{training data (X,y)}
\nomenclature[C,01]{\(D\)}{data distribution of dataset}
\nomenclature[C,02]{\(h\)}{classifier sampled from hypothesis space $\mathcal{H}$}
\nomenclature[C,03]{\(P_{t}\)}{prior classifier distribution over hypothesis space $\mathcal{H}$ at time $t$}
\nomenclature[C,03]{\(Q_{t}\)}{posterior classifier distribution over hypothesis space $\mathcal{H}$ at time $t$}
\nomenclature[C,04]{\(\ell\left(\cdot\right)\)}{bounded loss function}
\nomenclature[C,05]{\({\rm K}\)}{upper bound of the bounded loss $\ell\left(\cdot\right)$}
\nomenclature[C,06]{\({\rm KL}()\)}{Kullback-Leibler divergence}
\nomenclature[C,07]{\(\delta\)}{PAC-Bayes bound confidence probability}
\begin{multicols}{2}
\printnomenclature
\end{multicols}
\end{framed}
\end{table} 

\section{Preliminaries}
\label{section_Preliminaries}
This section introduces the GP/online GP algorithms and the online PAC-Bayes theory.

\subsection{GP algorithms}

{\bf Standard GP regression}
GP is a powerful non-parameter Bayesian algorithm, which can be used for regression, classification tasks. Standard GP model is entirely decided by its kernel function. From a function space perspective, the standard GP can be viewed as a distribution over function space with mean function $m$ and the covariance function $K$. The prior distribution $P$ in standard GP model is normally initialized as zero-mean Gaussian distribution $P(f)=\mathcal{N}(f \mid {\bf 0}, K)$, where $K$ (also $K(X,X)$) denotes the covariance matrix on the training dataset $X$. Thus the joint distribution of training data $X$ and testing data $X_\star$ can be calculated as

\begin{equation}
\left[\begin{array}{l}
f \\
f_{*}
\end{array}\right] \sim \mathcal{N} \left( \mathbf{0}, \left[ \begin{array}{cc}
K & K_{f\star} \\
K_{\star f} & K_{\star\star}
\end{array} \right] \right).
\end{equation}

Then the likelihood of training data $X$ can be derived as
\begin{equation}\label{GPR_likelihood}
P(y|f)=\mathcal{N}(f, \sigma_{y}^{2} {\rm I}).
\end{equation}


Furthermore, for some train dataset $(X, y)$, the predicted posterior distribution function value of $f_\star$ at the test data $X_\star$ can be derived from 
\begin{eqnarray}\label{GP_posterior}
\begin{aligned} 
Q(f_\star)= \mathcal{N} ( f \mid  & K_{f\star}^{\rm T} (K + \sigma_{y}^{2} {\rm I}  )^{-1} y, K_{\star\star} - K_{f\star}^{\rm T} (K + \sigma_{y}^{2} {\rm I} )^{-1} K_{f\star} ).
\end{aligned}
\end{eqnarray}

Different kernel functions can be selected in GP regression. The hyperparameters of such can be optimized by maximizing the {\it log marginal likelihood} $p(y | X, \theta)$
\begin{eqnarray}\label{GP_lml}
\begin{aligned} 
\log p(y | X, \theta) = & -\frac{1}{2} y^{\rm T} (K+\sigma_{y}^{2} {\rm I})^{-1} y -\frac{1}{2} \log |K+\sigma_{y}^{2} {\rm I} |  - \frac{n}{2} \log 2 \pi.
\end{aligned}
\end{eqnarray}

However, standard GPs cannot deal with streaming data, and hence several GP algorithms have been specifically designed for streaming settings.

{\bf Streaming GP algorithm}\label{stream_gp_review}
Data arrive sequentially in streaming settings, such that new data points $y_{new}$ are added to the previous dataset $y_{old}$  at each step. Our objective is to approximate the posterior distribution of the GP at each interval so it can be used for online prediction. Consider some newly arriving data $y_{new}$ the old approximate of posterior distribution $q_{\mathrm{old}}(f)$, which should be updated to the new approximation of posterior distribution $q_{\mathrm {new }}(f)$ (\cite{bui2017streaming})

\begin{equation}\label{stream_gp_q_old}
q_{\mathrm{old}}(f) \approx p\left(f \mid \mathbf{y}_{\mathrm{old}}\right)=\frac{1}{\mathcal{Z}_1\left(\theta_{\mathrm {old }}\right)} p\left(f \mid \theta_{\mathrm{old}}\right) p\left(\mathbf{y}_{\mathrm{old}} \mid f\right),
\end{equation}

\begin{equation}\label{stream_gp_q_new}
q_{\mathrm {new }}(f) \approx p\left(f \mid \mathbf{y}_{\mathrm {old }}, \mathbf{y}_{\mathrm {new }}\right)=\frac{1}{\mathcal{Z}_2\left(\theta_{\mathrm {new }}\right)} p\left(f \mid \theta_{\mathrm {new }}\right) p\left(\mathbf{y}_{\mathrm {old }} \mid f\right) p\left(\mathbf{y}_{\mathrm {new }} \mid f\right).
\end{equation}

The updated posterior distribution is then

\begin{equation}
\hat{p}\left(f \mid \mathbf{y}_{\mathrm {old }}, \mathbf{y}_{\mathrm {new }}\right)=\frac{\mathcal{Z}_1\left(\theta_{\mathrm {old }}\right)}{\mathcal{Z}_2\left(\theta_{\mathrm {new }}\right)} p\left(f \mid \theta_{\mathrm {new }}\right) p\left(\mathbf{y}_{\mathrm {new }} \mid f\right) \frac{q_{\mathrm {old }}(f)}{p\left(f \mid \theta_{\mathrm {old }}\right)} .
\end{equation}

However, the posterior distribution is intractable. Thus, the variational inference is used to update the kernel parameters and inducing points. Therefore, the optimization objective should be written as
\begin{equation}
\begin{aligned}
\operatorname{KL}\left[q_{\mathrm {new }}(f) \mid \hat{p}\left(f \mid \mathbf{y}_{\mathrm {old }}, \mathbf{y}_{\mathrm {new }}\right)\right] = \log \frac{\mathcal{Z}_2\left(\theta_{\mathrm {new }}\right)}{\mathcal{Z}_1\left(\theta_{\mathrm {old }}\right)}+\int \mathrm{d} f q_{\mathrm {new }}(f)\left[\log \frac{p\left(\mathbf{a} \mid \theta_{\mathrm {old }}\right) q_{\mathrm {new }}(\mathbf{b})}{p\left(\mathbf{b} \mid \theta_{\mathrm {new }}\right) q_{\mathrm {old }}(\mathbf{a}) p\left(\mathbf{y}_{\mathrm {new }} \mid f\right)}\right].
\end{aligned}
\end{equation}

\subsection{Online PAC-Bayes theorem}
Here, we introduce the online PAC-Bayes theory by \cite{haddouche2022online}.

In a standard supervised learning model, a set of $m$ dependent samples $S = \{z_i\}_{i=1}^m $ is randomly drawn from an unknown data distribution $\mathcal{D}$. Each sample $z_i = (x_i, y_i)$ consists of an input $x_i$ and its corresponding label $y_i$, where $x \in \mathcal{X}$ and $y \in \mathcal{Y}$. The learning objective is to find a classifier $h \in \mathcal{H}$ that predicts the label for new data $x^{\star}$, where $\mathcal{H}$ represents the hypothesis space and $\ell:\mathcal{H} \times \mathcal{Z} \to \mathbb{R}$ is the loss function to measure the quality of the classifier $h$.

An online PAC-Bayes framework must consider a sequence of \textit{randomized predictors}. First, we set a sequence of priors $P_i$ that start from a data-free distribution $P_1$. Then the sequence of posterior distribution $Q_i$ is developed where $Q_i = f(Q_1, Q_2, \cdot,Q_{i-1}, z_i)$. Here, only the loss function $\ell\left(\cdot\right)$ is considered, where $\ell\left(\cdot\right) < K$. Further, the prior distribution $P_i$ is independent of the data $z_i$ at time $i$. This leads to the following main result:

\begin{theorem} \label{online_pac_bayes}
(\cite{haddouche2022online}). Suppose that $P_i$ is the prior distribution over the hypothesis space $\mathcal{H}$, bounded loss function $\ell(h,z)$, $m$ instances $S$ sampled from data distribution $\mathcal{D}$, and confidence level $\delta \in (0,1]$. Then with a probability of at least $1 - \delta$, the following inequality holds for all posteriors distributions $Q_i \in \mathcal{M}$ 
\begin{equation}
\sum_{i=1}^m \mathbb{E}_{h_i \sim Q_i}\left[\mathbb{E}\left[\ell\left(h_i, z_i\right) \right]\right] \leq \sum_{i=1}^m \mathbb{E}_{h_i \sim Q_i}\left[\ell\left(h_i, z_i\right)\right]+\frac{\operatorname{KL}\left(Q_i \mid P_i\right)}{\lambda}+\frac{\lambda m K^2}{2}+\frac{\log (1 / \delta)}{\lambda}
\end{equation}
\end{theorem} 

The left-hand side of the bound denotes the sum of the average expected loss, which is similar to {\it generalization error}. The right side of the bound consists of three items: the sum of the empirical loss, a regularization item, and the constant item. The regularization item involves the distance between the prior distribution $P_i$ and the posterior distribution $Q_i$, which avoids overfitting. Unlike the traditional PAC-Bayes theory by \cite{mcallester1999some}, which is compatible with batch data, the online PAC-Bayes theorem provides a framework for online predictions with a guarantee of generalized performance.

Furthermore, the online PAC-Bayes theory motivates an {\it online PAC-Bayes training bound} (\ref{online_pac_bayes_train}) and an {\it online PAC-Bayes testing bound} (\ref{online_pac_bayes_test}).


\begin{corollary}(OPB Train)
\label{online_pac_bayes_train}
\begin{equation}
\sum_{i=1}^m \mathbb{E}_{h_i \sim \hat{Q}_{i+1}}\left[\mathbb{E}\left[\ell\left(h_i, z_i\right) \right]\right] \leq \sum_{i=1}^m \mathbb{E}_{h_i \sim \hat{Q}_{i+1}}\left[\ell\left(h_i, z_i\right)\right] + \frac{\operatorname{KL}\left(\hat{Q}_{i+1} \mid P_i\right)}{\lambda}+\frac{\lambda m K^2}{2}+\frac{\log (1 / \delta)}{\lambda}.
\end{equation}
\end{corollary}

\begin{corollary}(OPB Test)
\label{online_pac_bayes_test}
\begin{equation}
\sum_{i=1}^m \mathbb{E}_{h_i \sim \hat{Q}_i}\left[\mathbb{E}\left[\ell\left(h_i, z_i\right) \right]\right] \leq \sum_{i=1}^m \mathbb{E}_{h_i \sim \hat{Q}_i}\left[\ell\left(h_i, z_i\right)\right]+\frac{\lambda m K^2}{2}+\frac{\log (1 / \delta)}{\lambda}.
\end{equation}
\end{corollary}

Here, the {\it online PAC-Bayes training bound} (\ref{online_pac_bayes_train}) is used to optimize the GP or neural network, and an {\it online PAC-Bayes testing bound} (\ref{online_pac_bayes_test}) is used to quantify the prediction’s accuracy.

\section{Online PAC-Bayes Gaussian process algorithm}
\label{section_online_pacgp}
\begin{figure}[tbp]
\centering 
\includegraphics[width=0.8\textwidth]{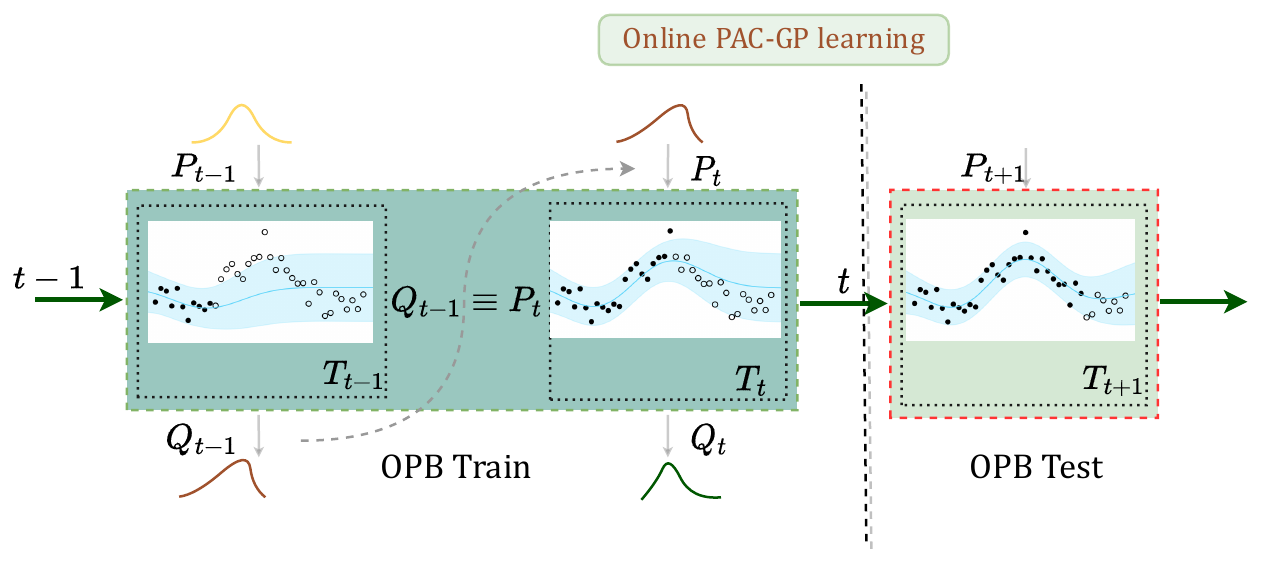}
\caption{The online PACGP framework. When new observed data $z_t$ arrive sequentially at time $t$, then the posterior distribution $Q_t$ can be updated based on the prior distribution $P_t$ and a new observation $z_t$. Then, at time $z_{t+1}$, the previous posterior distribution $Q_t$ is considered to be the prior distribution $P_{t+1}$.}
\label{online_PACGP_frame}
\end{figure}

To achieve a balance between the generalization error upper bound and accuracy, in this work, the online PAC-Bayes GP framework is proposed based on the online GP algorithm \ref{stream_gp_review} and the online PAC-Bayes theory \ref{online_pac_bayes}. As shown in Fig. \ref{online_PACGP_frame}, when new observed data $z_t$ arrive sequentially at time $t$, then the posterior distribution $Q_t$ is updated based on the prior distribution $P_t$ and the new observation $z_t$. Then, at time $z_{t+1}$, the previous posterior distribution $Q_t$ is considered to be the prior distribution $P_{t+1}$. Using the {\it online PAC-Bayes training bound} and the {\it online PAC-Bayes testing bound}, the training and prediction procedure of online GP algorithms are developed, and the hyperparameter and inducing points are updated in an online fashion. 

\subsection{Learning online GP}
The PAC-Bayes theory provides a numerical performance guarantee with data of an unknown distribution. One can optimize their algorithms by directly selecting the PAC-Bayes bound as the training objective. Here, based on the {\it online PAC-Bayes training bound}, the optimization objective can be rewritten as 
\begin{equation}\label{online_pacgp_obj}
\begin{aligned}
J = \sum_{i=1}^m \mathbb{E}_{h_i \sim \hat{Q}_{i+1}}\left[\ell\left(h_i, y_i\right)\right] + \frac{\operatorname{KL}\left(\hat{Q}_{i+1} \mid P_i\right)}{\lambda}+\frac{\lambda m K^2}{2}+\frac{\log (1 / \delta)}{\lambda},
\end{aligned}
\end{equation}

This optimization objective consists of three items: an empirical loss, a regularization item, and a constant item.

{\bf Calculating the KL divergence} The KL divergence involves the distance between the prior distribution and posterior distributions. Given a streaming GP algorithm, one can approximate the KL divergence from its prior distribution $q_{\mathrm{old}}$ and the posterior distribution $q_{\mathrm{new}}$ at time $t$
\begin{equation}\label{online_PACgp_kl}
\begin{aligned}
{\rm KL}(Q\mid P) & = {\rm KL}\left( \hat{p}(f \mid \mathbf{y}_{\mathrm {old }}, \mathbf{y}_{\mathrm {new }}) \mid q_{\mathrm{old}}(f) \right) \\
& \approx {\rm KL}(q_{\mathrm{new}}(f)\mid q_{\mathrm{old}}(f)).
\end{aligned}
\end{equation}

Here, the prior distribution $q_{\mathrm{old}}$ and the posterior distribution $q_{\mathrm{new}}$is inferred in (\ref{stream_gp_q_old}) and (\ref{stream_gp_q_new}), respectively.

{\bf Calculating the Bounded loss function}
The online PAC-Bayes bound \ref{online_pac_bayes} is only available for the bounded loss function. In the online PACGP framework, the bounded loss function is still chosen as follows
\begin{equation}\label{bounded_loss_exp}
\ell_{\exp }(y, \widehat{y})=1-\exp \left(-((y-\widehat{y}) / \varepsilon)^{2}\right),
\end{equation}
where the $\epsilon > 0$ denotes a scale parameter. 

Assuming that the posterior distribution $Q_i \sim \mathcal{N}(m_i, \sigma_i^2)$, then the expectation item $\sum_{i=1}^m  \mathbb{E}_{h_i \sim \hat{Q}_{i}}\left[ \ell(h_i, y_i)\right]$ equals 

\begin{equation}\label{online_PACgp_bounded_loss}
\begin{aligned}
\sum_{i=1}^m \mathbb{E}_{h_i \sim \hat{Q}_{i}}\left[ \ell(h_i, y_i) \right] &= \sum_{i=1}^N \mathbb{E}_{h \sim \hat{Q}_{i}} \left[ \ell(h, y_i) \right] \\
&= \sum_{i=1}^N \int {\rm d}h \mathcal{N} ( v \mid m_i, \sigma_i^2) \ell (h, y_i),
\end{aligned}
\end{equation}
where the integral part can be derived as
\begin{equation}
\int d h \mathcal{N}(v \mid m_i, \sigma_i^2) \ell_{\exp} (y_i, h) = 1 - \frac{1}{\sqrt{ 1+\frac{2  \sigma_i^2}{\varepsilon^2}}} \exp \left[ - \frac{(y_i-m_i)^2}{2 \sigma_i + \varepsilon^2} \right].
\end{equation}

For other choices of bounded loss functions, please see Appendix \ref{app_Bounded_loss_func}.

As discussed before, the sum of the expected bounded loss function and the KL divergence are given in (\ref{online_PACgp_bounded_loss}) and (\ref{online_PACgp_kl}). The constant only relates to the constant upper bound $K$ of the loss function and the hyperparameter confidence probability $\delta$. Therefore the optimization objective can be calculated by substituting (\ref{online_PACgp_bounded_loss}) and (\ref{online_PACgp_kl}) into (\ref{online_pacgp_obj}). The specific pseudo code is shown in Algorithm \ref{pseduo_alg_online_pacgp}.

\subsection{Onine GP inference}
In the prediction stage, utilizing the online PAC-Bayes theory, the prediction performance can be measured via
\begin{equation}\label{online_pacgp_testing}
\begin{aligned}
J_{test} = \sum_{i=1}^m \mathbb{E}_{h_i \sim \hat{Q}_i}\left[\ell(h_i, z_i)\right] + \frac{\lambda m K^2}{2} + \frac{\log (1 / \delta)}{\lambda},
\end{aligned}
\end{equation}
This upper bound quantifies how efficient will the learned model predictions be in the streaming setting.

Compare with other online GP algorithms, streaming GP and WISKI approaches aim to minimize the estimation error by optimizing the marginal log-likelihood. However, the online PACGP algorithm tries to directly optimize the generalization error upper bound, which can achieve generalization performance guarantee.

\begin{table*}[tbp]
\centering
\caption{Comparison of different online GP algorithms. The streaming GP and WISKI approaches aim to minimize the estimation error by optimizing the marginal log-likelihood. The online PACGP algorithm tries to directly optimize the generalization error upper bound.}
\label{dif_obj_onine_GP_PACGP}
\begin{tabular}{ccc}
\toprule[1pt]
         & Objective           &    Optimization objective           \\ \hline \\
\vspace{10pt}
Streaming GP       &  Minimize error     &    $\operatorname{KL}\left[q_{\mathrm {new }}(f) \mid \hat{p}\left(f \mid \mathbf{y}_{\mathrm {old }}, \mathbf{y}_{\mathrm {new }}\right)\right]$ \\
\vspace{10pt}
WISKI     &  Minimize error     &  $\log p(y \mid X,\theta)$ \\
\vspace{6pt}
Online PACGP &  Minimize error and generalization  bound  & $\sum_{i=1}^m \mathbb{E}_{h_i \sim \hat{Q}_{i+1}}\left[\ell\left(h_i, z_i\right)\right] + \frac{\operatorname{KL}\left(\hat{Q}_{i+1} \mid P_i\right)}{\lambda}+Cons$ \\
\toprule[1pt]
\end{tabular}
\end{table*}

\begin{algorithm}[tbp]
\caption{Online PACGP algorithm}
\label{pseduo_alg_online_pacgp}
  \begin{algorithmic}
    \Require
        new observed samples $(X_t, y_t)$
    \Ensure
        Learned parameters: including kernel parameters $\theta_{new}$ and inducing points $z_{new}$.
    \State Initializing prior distribution ${P_1}$,  kernel parameters $\theta$ and inducing points $z$;
    \While{not done}
            \State Calculating KL divergence between prior and  posterior distributions via (\ref{online_PACgp_kl}) 
            \State Calculating the sum of expected bounded loss function via (\ref{online_PACgp_bounded_loss}) 
            \State Calculating PAC-Bayes objective $J$ via (\ref{online_pacgp_obj}) as optimization function
            \State Minimizing the PAC-Bayes objective $J$
    \EndWhile\\ 
    \Return $\theta_{new}$ and $z_{new}$;
  \end{algorithmic}
\end{algorithm}

\section{Experiments}\label{section_experiments}
In this part, we first verify the performance of our algorithm on several regression datasets. We then outline the experiments designed to examine the behavior of the parameters.

\subsection{Experimental settings}

{\bf Baselines}
The baseline algorithms selected for comparison included
\begin{enumerate}
    \item O-SGPR \cite{bui2017streaming}. O-SGPR was developed from sparse GP regression algorithms. This algorithm can deploy probabilistic GP models in a streaming setting. The advantage of this algorithm is that it can provide explicit expressions of the posterior distribution. However, O-SGPR is only suitable for Gaussian likelihood functions.
    \item O-SVGP \cite{bui2017streaming}. O-SVGP performs streaming sparse GP approximation through variational inference. Unlike O-SGPR, O-SVGP is applicable to non-Gaussian likelihoods, but its limitation is that this approach sacrifices closed-form expressions of the posterior distribution.
    \item WISKI \cite{stanton2021kernel}. Based on the structured kernel interpolation, WISKI can deliver constant-time O(1) online GP updates with respect to the number of newly arrived data while retaining exact inference. Compared to O-SGPR and O-SVGP, which with exact inference, this algorithm also offers competitive performance.
\end{enumerate}

{\bf Datasets} {We verified the proposed online PAC-Bayes GP approach using three datasets.}

\begin{enumerate}
    \item \textit{Synthetic datasets}. To verify the effectiveness of the proposed online PACGP approach, two synthetic datasets, including \textit{Sinusoidal dataset} and \textit{Cosinusoidal dataset}. The \textit{sinusoidal} is an artificial dataset, generated by $sin$ function: $y = \sin(4x)$, and the \textit{cosinusoidal} dataset is generated by $cos$ function: $y = \cos(4x)$.
    \item {\textit{Stock price dataset}.  The stock price dataset is a low-dimensional dataset containing of a total of 251 cases, with 13 attributes in each case of the dataset. This dataset is available at \url{https://raw.githubusercontent.com/trungngv/cogp/master/data/fx/fx2007-processed.csv}}.
    
    In addition, for each dataset data arrive sequentially for both training and testing. To simplify the technique we also normalize all features and outputs of those datasets to mean zero and unit variance.
\end{enumerate}

{\bf Parameter setting}
For the O-SGPR and O-SVGP, the 
We used the same number of inducing points for O-SGPR, O-SVGP, WISKI, and online PACGP. The implementation of these four algorithms is under the framework of GpyTorch \cite{gardner2018gpytorch}. The Kernel function is selected as a spectral mixture kernel. The gradient optimization algorithm is Adam \cite{kingma2014adam}. The learning rate is set as $1e-1$ for likelihood and kernel parameters and $1e-2$ for variational parameters. The inducing points are initialized by linearly spaced points. The models were pre-trained on $5\%$ of the training examples and then trained online for the remaining $95\%$. When the new data points arrive, we update with a single optimization step for each corresponding method. For the online PACGP model, the model is pre-trained as same as O-SGPR, the $\epsilon^2$ in bounded loss function is selected as 0.01, and confidence probability $\lambda$ is set as $\frac{1}{len(x)}$, where the number of current arrived data $x$. The code is available at  \url{https://github.com/tyliu22/online_pacgp}.

\subsection{experiment results}
In this part, we verify the effectiveness of the proposed algorithm.

\begin{figure*}[tbp]
	\centering
     \subfigure[Online PACGP for {\it sin} with non-iid data]{
	    \label{Online_PACGP_syn_data_sin_non_iid}
        \includegraphics[width=0.8\textwidth]{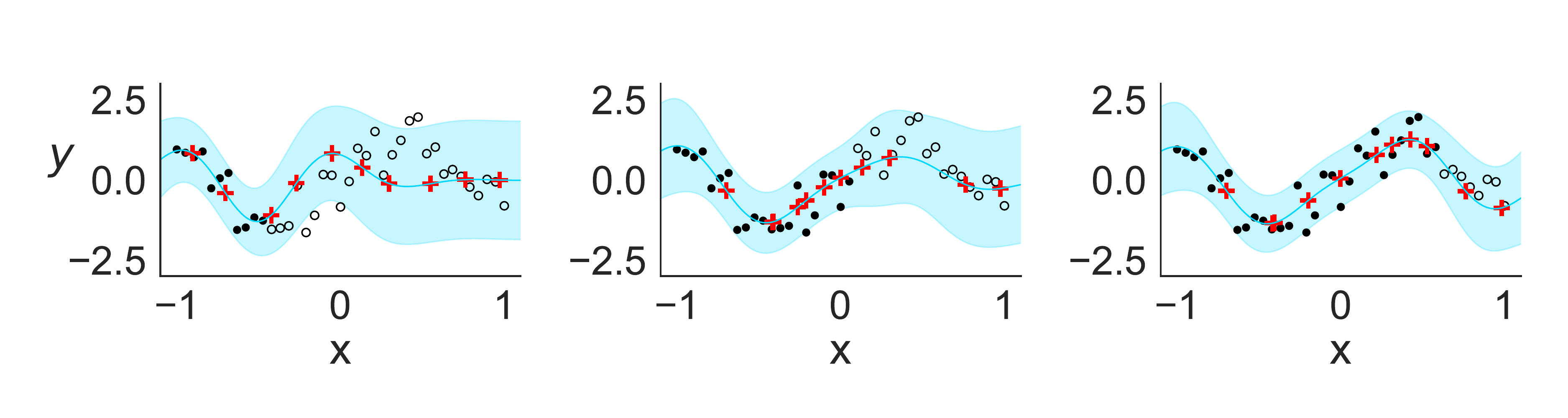}
        }
        \quad
	\subfigure[Online PACGP for {\it sin} with iid data]{
        \label{Online_PACGP_syn_data_sin_iid}
        \includegraphics[width=0.8\textwidth]{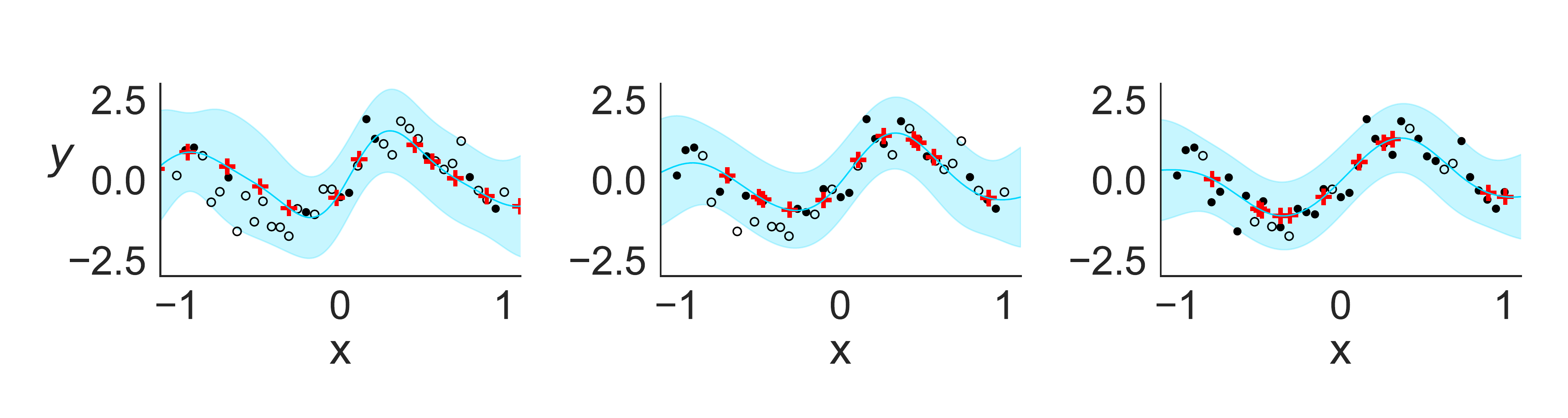}
        }
        \quad
    \subfigure[Online PACGP for {\it cos} with non-iid  data]{
	    \label{Online_PACGP_syn_data_cos_non_iid}
        \includegraphics[width=0.8\textwidth]{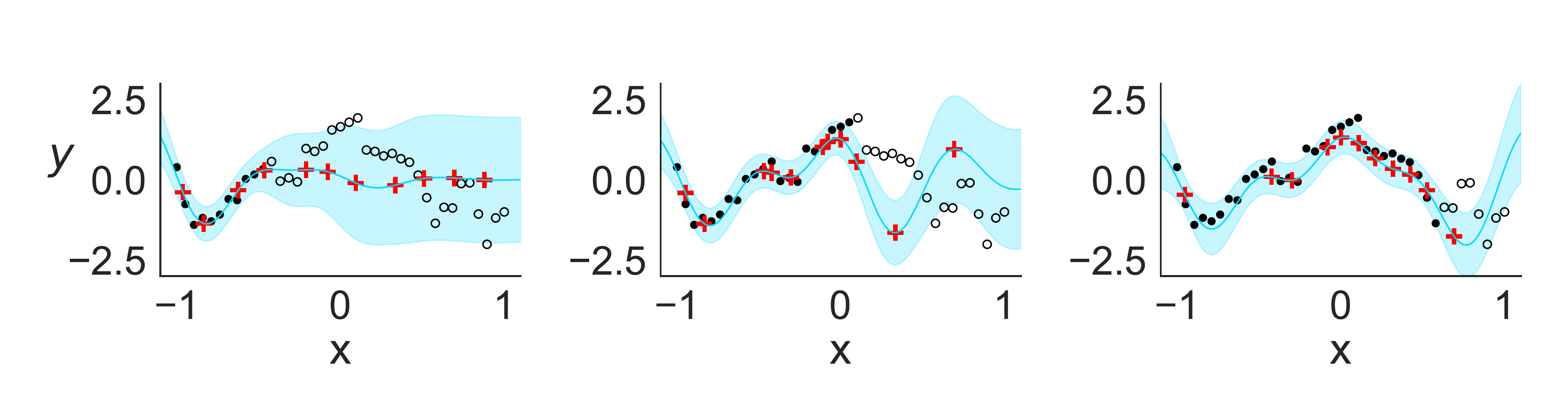}
        }
        \quad
	\subfigure[Online PACGP for {\it cos} with iid data]{
        \label{Online_PACGP_syn_data_cos_iid}
        \includegraphics[width=0.8\textwidth]{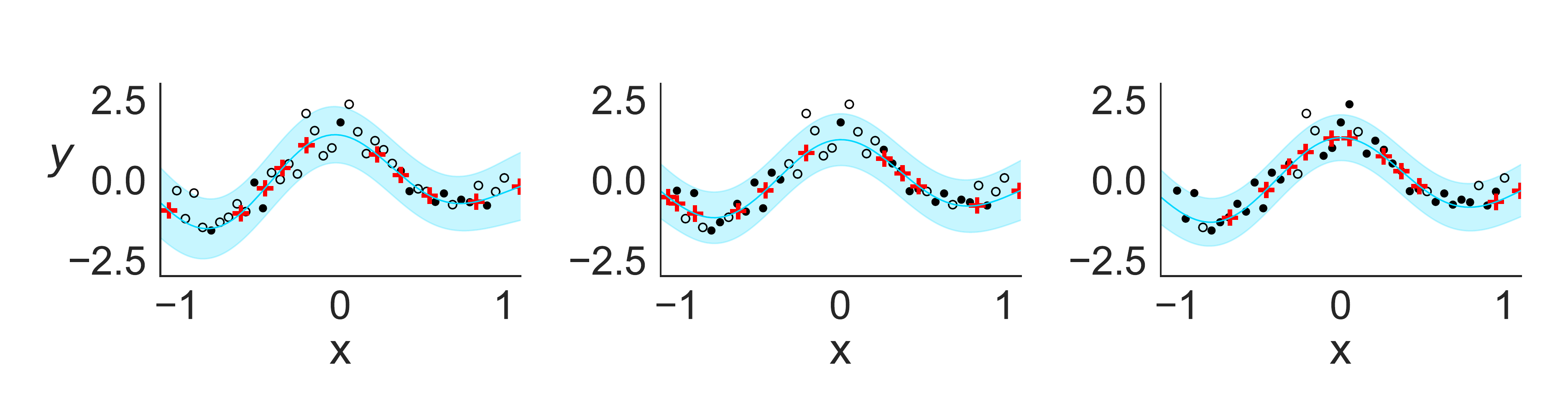}
        } 
	\caption{
	Effectiveness of online PACGP approach. Here, non-iid data means the model is trained on observations in a time series fashion, and iid data means the model is trained on observations in a randomly ordered fashion. Red points denote the inducing points, black points denote the training data, and the circle represents the testing points. Obviously, the proposed online PACGP approach can update the hyperparameter and inducing points in the online fashion, and also achieve good estimation performance.}
	\label{Online_PACGP_for_Synthetic_data}
\end{figure*}

{\bf Synthetic model}
We verify the effectiveness of the proposed online PAC-Bayes GP approach on two synthetic datasets, including \textit{Sinusoidal dataset} and \textit{Cosinusoidal dataset}. Here we consider two settings: non-iid data means the model is trained on observations in a time series fashion, and iid data means the model is trained on observations in a randomly ordered fashion. As shown in Fig. \ref{Online_PACGP_for_Synthetic_data}, the experiment results demonstrate that the proposed online PAC-Bayes GP approach can update the hyperparameter and inducing points  in the online fashion, and also achieve good estimation performance.

\begin{figure*}[tbp]
	\centering
     \subfigure[O-SGPR for stock price with non-iid data]{
	    \label{Rep_GPR_post_fit_paras}
        \includegraphics[width=0.8\textwidth]{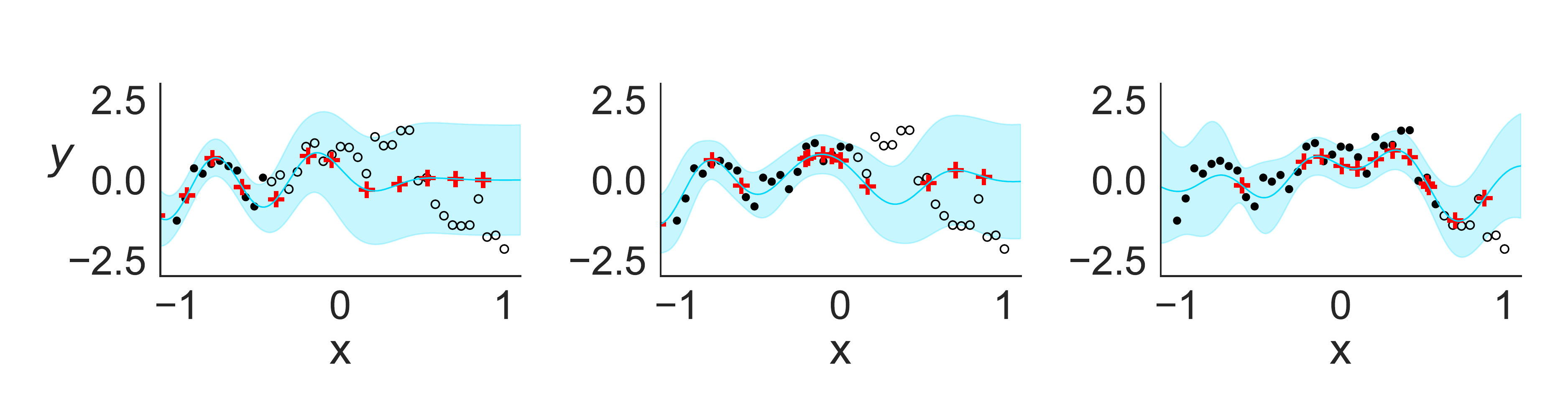}
        }
        \quad
	\subfigure[O-SVGP for stock price with non-iid data]{
        \label{Rep_NIGPR_post_fit_paras}
        \includegraphics[width=0.8\textwidth]{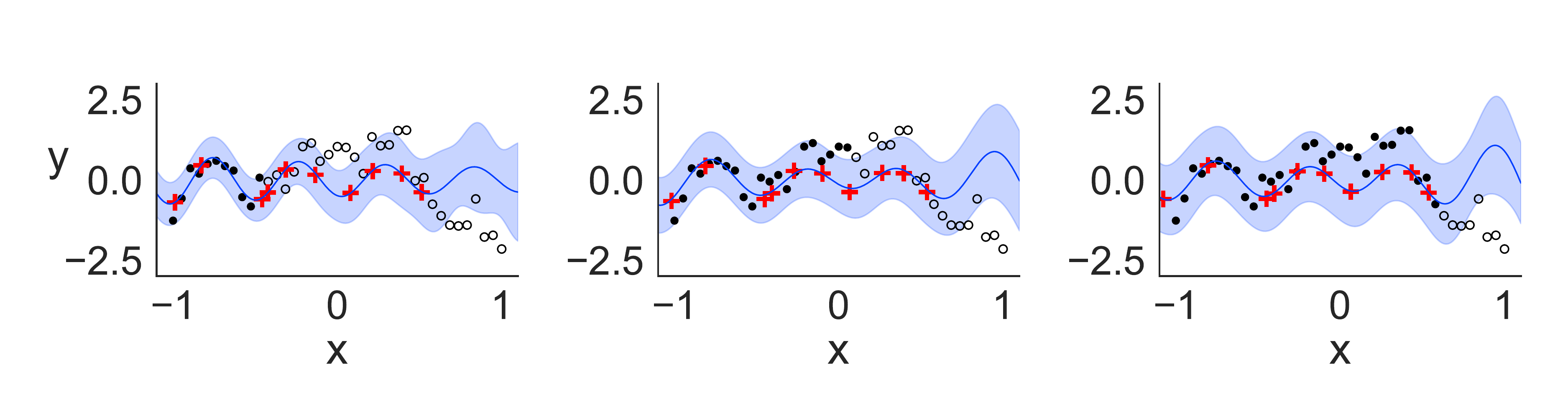}
        }
        \quad
    \subfigure[WISKI for stock price with non-iid data]{
	    \label{Rep_GPR_post_init_paras}
        \includegraphics[width=0.8\textwidth]{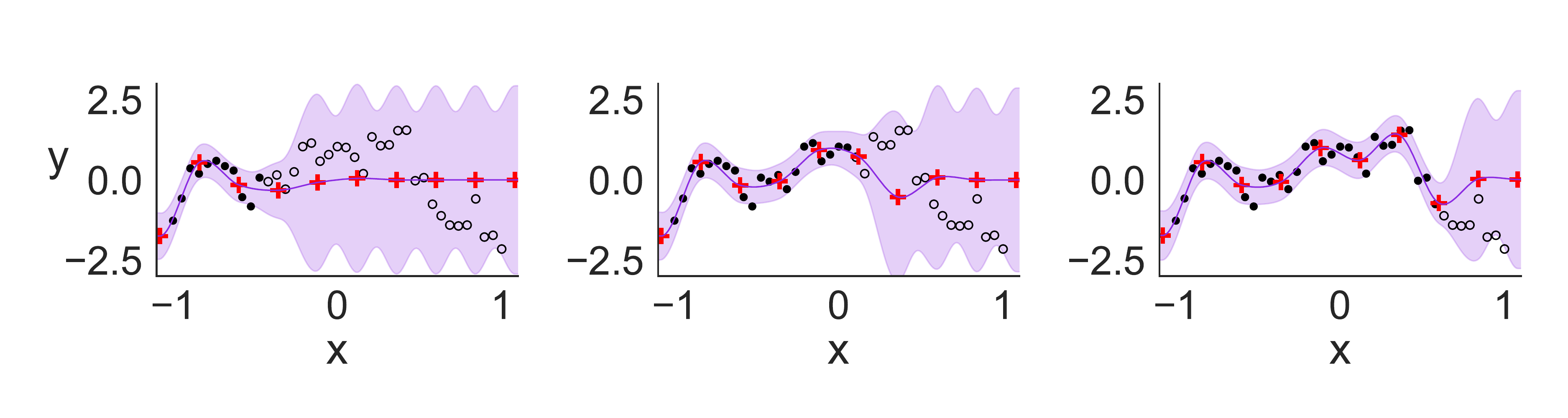}
        }
        \quad
	\subfigure[Online PACGP for stock price with non-iid data]{
        \label{Rep_NIGPR_post_init_paras}
        \includegraphics[width=0.8\textwidth]{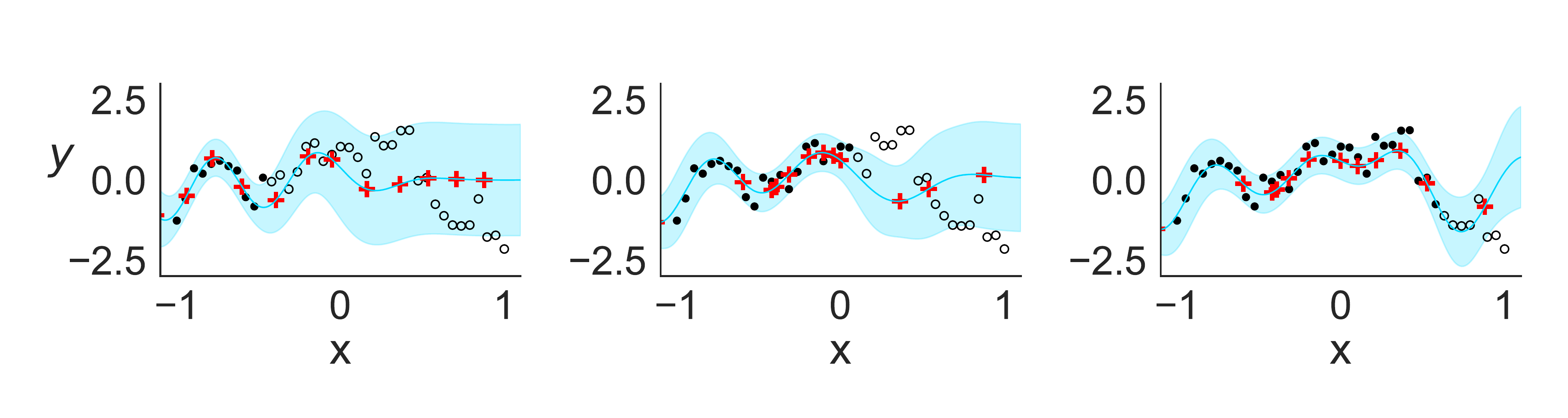}
        } 
	\caption{A comparison of four online GP approaches with the non-iid dataset, such as O-SGPR, O-SVGP, WISKI, and Online PACGP. Here, non-iid data means the model is trained on observations in a time series fashion. Red points denote the inducing points, black points denote the training data, and the circle represents the testing points. Compared with other algorithms, the proposed Online PACGP achieves better estimation performance for time series data in the online fashion.}
	\label{Online_PACGP_for_stock_price_non_iid_data}
\end{figure*}

\begin{figure*}[tbp]
	\centering
     \subfigure[O-SGPR for stock price with iid data]{
	    \label{Rep_GPR_post_fit_paras}
        \includegraphics[width=0.8\textwidth]{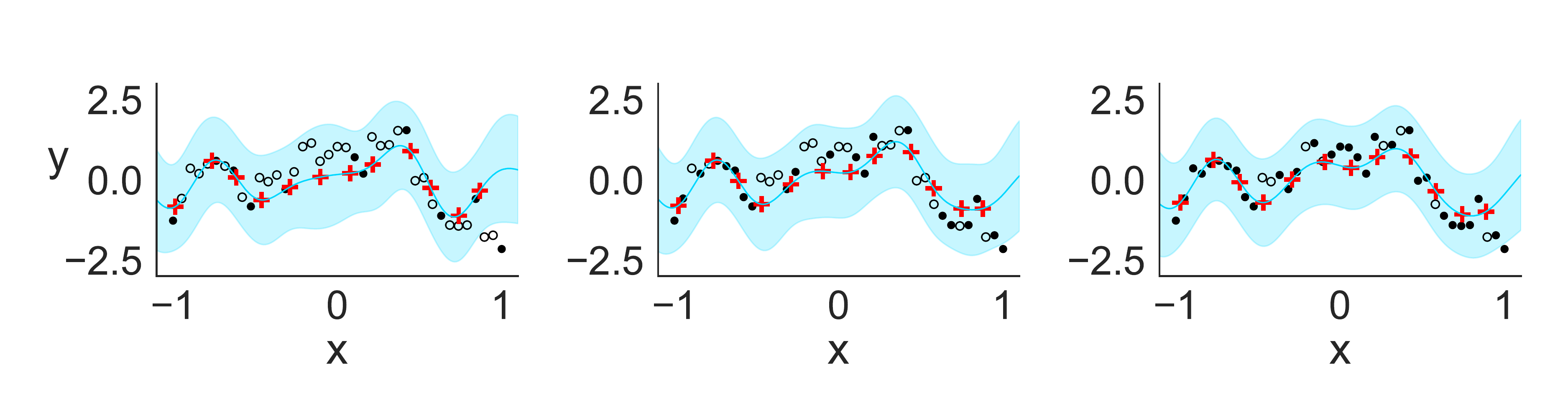}
        }
        \quad
	\subfigure[O-SVGP for for stock price with iid data]{
        \label{Rep_NIGPR_post_fit_paras}
        \includegraphics[width=0.8\textwidth]{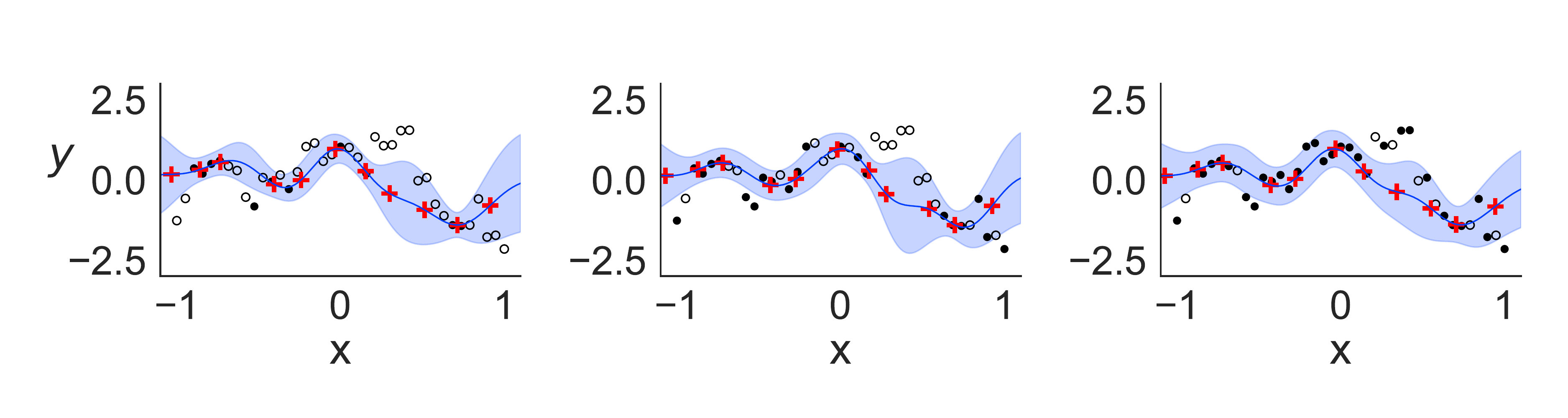}
        }
        \quad
    \subfigure[WISKI for stock price with iid data]{
	    \label{Rep_GPR_post_init_paras}
        \includegraphics[width=0.8\textwidth]{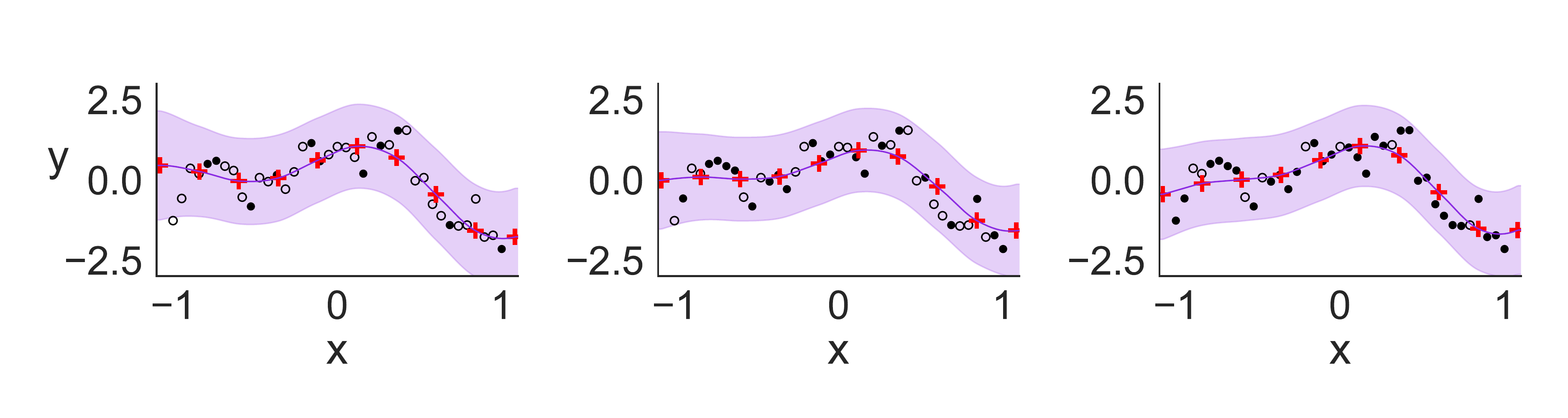}
        }
        \quad
	\subfigure[Online PACGP for stock price with iid data]{
        \label{Rep_NIGPR_post_init_paras}
        \includegraphics[width=0.8\textwidth]{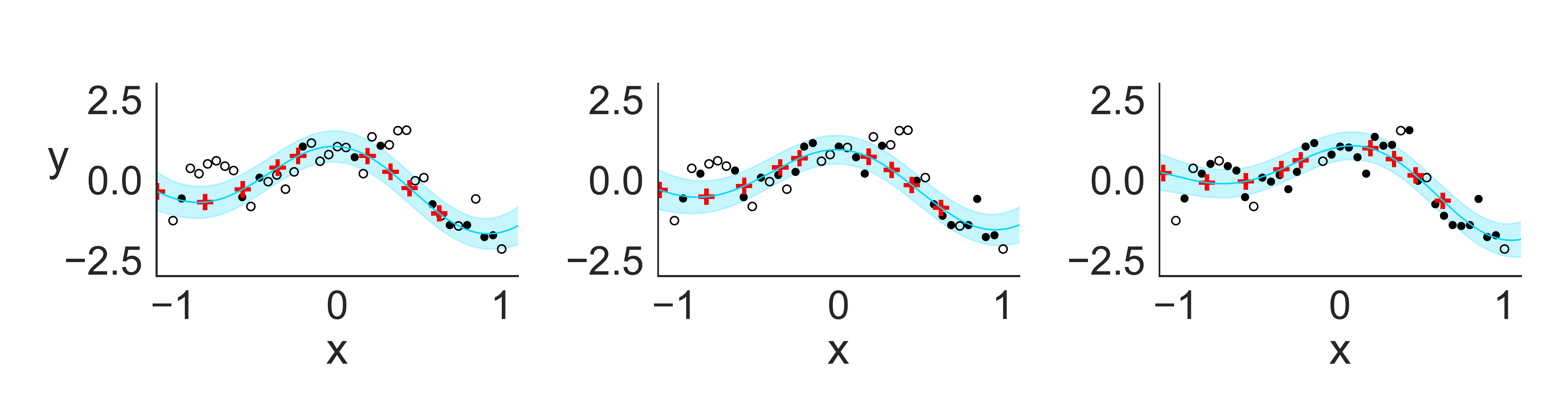}
        } 
	\caption{A comparison of four online GP approaches with the iid dataset, such as O-SGPR, O-SVGP, WISKI, and Online PACGP. Here, iid data means the model is trained on observations in a randomly ordered fashion. Red points denote the inducing points, black points denote the training data, and the circle represents the testing points.  Compared with other algorithms, the proposed Online PACGP also can achieve competitive estimation performance.}
	\label{Online_PACGP_for_stock_price_iid_data}
\end{figure*}

{\bf  Regression dataset}
We compare four online GP approaches, such as O-SGPR, O-SVGP, WISKI, and Online PACGP, with both the non-iid dataset and iid dataset. As shown in Fig. \ref{Online_PACGP_for_stock_price_non_iid_data} and Fig. \ref{Online_PACGP_for_stock_price_iid_data}, all those four online GP approaches can achieve online hyper-parameter learning and pseudo-point location optimization. Besides, as shown in Table. \ref{Tab_Online_PACGP_for_stock_price}, compared with other algorithms, the proposed Online PACGP achieve competitive estimation performance.

\begin{table}[]
\centering
\caption{A comparison of four online GP approaches, such as O-SGPR, O-SVGP, WISKI, and Online PACGP,  with both the non-iid dataset and iid dataset. The averaged train MSE and test MSE are analyzed. Compared with other algorithms, the proposed Online PACGP achieve competitive estimation performance.}
\label{Tab_Online_PACGP_for_stock_price}
\begin{tabular}{llllllll}
\hline
\toprule[1pt]
\multicolumn{2}{l}{\multirow{2}{*}{}}       & \multicolumn{2}{l}{$t=10$} & \multicolumn{2}{l}{$t=20$} & \multicolumn{2}{l}{$t=30$} \\  \cline{3-4}  \cline{5-6}  \cline{7-8} 
\multicolumn{2}{l}{}                        & train error  & test error    & train error      & test error     & train error      & test error      \\ \hline
\multirow{4}{*}{IID data}    & O-SGPR       & 0.3965       & 0.4834  & 0.2784     & {\bf 0.4120}    & 0.2484     & {\bf 0.3537}   \\
                             & O-SVGP       & 0.0835       & {\bf 0.4322}  & 0.1936     & 0.4753    & 0.2439     & 0.4938    \\
                             & WISKI        & 1.1749       & 1.4999  & 1.1428     & 1.7391    & 1.6884     & 1.4223    \\
                             & Online PACGP & {\bf 0.0257}       & 1.0146  & {\bf 0.0637}     & 1.1384    & {\bf 0.1363}     & 1.1595    \\ \hline
\multirow{4}{*}{No-IID data} & O-SGPR       & 0.1147       & {\bf 1.2327}  & 0.1093     & 2.3115    & 0.2748     & {\bf 1.3628}    \\
                             & O-SVGP       & {\bf 0.1048}       & 1.2958  & 0.3194     & 2.0988    & 0.4395     & 3.9509    \\
                             & WISKI        & 0.6075       & 1.2334  & 0.7099     & {\bf 1.6777}    & 0.8696     & 1.8340    \\
                             & Online PACGP & 0.1146       & 1.2329  & {\bf 0.1035}     & 2.2901    & {\bf 0.2632}     & 1.5539    \\ 
                             \cline{1-8} 
                             \toprule[1pt]
\end{tabular}
\end{table}

\section{Conclusion and future work}\label{section_conclusion}
In this paper, we proposed a novel online PAC-Bayes GP framework to achieve a quantified guarantee of generalization performance in the online fashion. Furthermore, an online PAC-Bayes GP algorithm with a bounded loss function is developed to offer a balance between the generalization error upper bound and accuracy. Experiments illustrate the effectiveness of the proposed algorithm. Compared with other algorithms, the proposed online PACGP method enables a numerical generalization performance guarantee with competitive accuracy performance. In the future, we will focus on applying the proposed algorithm to classification, Bayesian optimization, and active learning settings.

\section*{Acknowledgment}
This work was supported by the Australian Research Council through the Discovery Project under Grant DP200100700. Tianyu Liu is supported by the International Research Training Program Scholarship (IRTP) of Australia.

\addcontentsline{toc}{section}{References}
\bibliographystyle{apalike}
\bibliography{references}

\section*{Appendix}

\subsection*{Bounded loss functions}\label{app_Bounded_loss_func}
Following the Ref \cite{reeb2018learning}, the bounded loss functions $\ell(\cdot)$ and those integral parts can be derived as 
\begin{equation}
\begin{aligned}
\ell_{\mathbb{1}}(y, \widehat{y}) &=\mathbb{1}_{|y-\widehat{y}|>\varepsilon}=\mathbb{1}_{\widehat{y} \notin[y-\varepsilon, y+\varepsilon]}, \\
\ell_2(y, \widehat{y}) &=\min \left\{((y-\widehat{y}) / \varepsilon)^2, 1\right\}, \\
\ell_{\exp }(y, \widehat{y}) &=1-\exp \left[-((y-\widehat{y}) / \varepsilon)^2\right], \\
\ell_{\pm}(y, \widehat{y}) &=\mathbb{1}_{\widehat{y} \notin\left[r_{-}(y), r_{+}(y)\right]}.
\end{aligned}
\end{equation}

Assuming that the posterior distribution $Q_i \sim \mathcal{N}(m_i, \sigma_i^2)$, then the expectation item $\sum_{i=1}^m  \mathbb{E}_{h_i \sim \hat{Q}_{i}}\left[ \ell(h_i, y_i)\right]$ equals 

\begin{equation}\label{online_PACgp_bounded_loss}
\begin{aligned}
\sum_{i=1}^m \mathbb{E}_{h_i \sim \hat{Q}_{i}}\left[ \ell(h_i, y_i) \right] &= \sum_{i=1}^N \mathbb{E}_{h \sim \hat{Q}_{i}} \left[ \ell(h, y_i) \right] \\
&= \sum_{i=1}^N \int {\rm d}h \mathcal{N} ( v \mid m_i, \sigma_i^2) \ell (h, y_i),
\end{aligned}
\end{equation}
where the integral part can be derived as
\begin{equation}
\int d h \mathcal{N}(v \mid m_i, \sigma_i^2) \ell_{\exp} (y_i, h) = 1 - \frac{1}{\sqrt{ 1+\frac{2  \sigma_i^2}{\varepsilon^2}}} \exp \left[ - \frac{(y_i-m_i)^2}{2 \sigma_i + \varepsilon^2} \right].
\end{equation}

For the other three bounded loss functions, the integral part can be calculated as 
\begin{equation}
\int d v \mathcal{N}\left(v \mid \widehat{m}_i, \widehat{\sigma}_i^2\right) \ell_{\mathbb{1}}\left(y_i, v\right)=\Phi\left(\frac{y_i-\varepsilon-\widehat{m}_i}{\widehat{\sigma}_i}\right)+1-\Phi\left(\frac{y_i+\varepsilon-\widehat{m}_i}{\widehat{\sigma}_i}\right),
\end{equation}

\begin{equation}
\begin{aligned}
\int d v \mathcal{N}\left(v \mid \widehat{m}_i, \widehat{\sigma}_i^2\right) \ell_2\left(y_i, v\right)=&\left(1-\frac{\left(y_i-\widehat{m}_i\right)^2+\widehat{\sigma}_i^2}{\varepsilon^2}\right)\left(\Phi\left(\frac{y_i-\varepsilon-\widehat{m}_i}{\widehat{\sigma}_i}\right)-\Phi\left(\frac{y_i+\varepsilon-\widehat{m}_i}{\widehat{\sigma}_i}\right)\right) \\
+1 &-\frac{\widehat{\sigma}_i}{\sqrt{2 \pi} \varepsilon^2}\left(y_i-\varepsilon-\widehat{m}_i\right) e^{-\left(y_i+\varepsilon-\widehat{m}_i\right)^2 /\left(2 \widehat{\sigma}_i^2\right)} \\
&-\frac{\widehat{\sigma}_i}{\sqrt{2 \pi} \varepsilon^2}\left(y_i+\varepsilon-\widehat{m}_i\right) e^{-\left(y_i-\varepsilon-\widehat{m}_i\right)^2 /\left(2 \widehat{\sigma}_i^2\right)},
\end{aligned}
\end{equation}


\begin{equation}
\int d v \mathcal{N}\left(v \mid \widehat{m}_i, \widehat{\sigma}_i^2\right) \ell_{\pm}\left(y_i, v\right)=\Phi\left(\frac{r_{-}\left(y_i\right)-\widehat{m}_i}{\widehat{\sigma}_i}\right)+1-\Phi\left(\frac{r_{+}\left(y_i\right)-\widehat{m}_i}{\widehat{\sigma}_i}\right),
\end{equation}
 where the cumulative distribution function (CDF) for the normal distribution is defined as
\begin{equation}
\Phi(z):=\int_{-\infty}^z \frac{1}{\sqrt{2 \pi}} e^{-t^2 / 2} {\rm d}t.
\end{equation}



\end{document}